\documentclass{bmvc2k}

\title{Solving Visual Madlibs with Multiple Cues}

\addauthor{Tatiana Tommasi}{ttommasi@cs.unc.edu}{1}
\addauthor{Arun Mallya}{amallya2@illinois.edu}{2}
\addauthor{Bryan Plummer}{bplumme2@illinois.edu}{2}
\addauthor{Svetlana Lazebnik}{slazebni@illinois.edu}{2}
\addauthor{Alexander C. Berg}{aberg@cs.unc.edu}{1}
\addauthor{Tamara L. Berg}{tlberg@cs.unc.edu}{1}

\addinstitution{
 University of North Carolina at\\
 Chapel Hill, (NC) USA
}
\addinstitution{
 University of Illinois at\\ 
 Urbana-Champaign, (IL) USA
}

\runninghead{T. Tommasi \bmvaEtAl}{Solving Visual Madlibs with Multiple Cues}

\def\eg{\emph{e.g}\bmvaOneDot}
\def\ie{\emph{i.e}\bmvaOneDot}

\def\etc{\emph{etc}\bmvaOneDot}

\usepackage{multirow}

\begin{document}

\maketitle

\begin{abstract}

This paper focuses on answering fill-in-the-blank style multiple choice questions from the Visual Madlibs dataset. 
Previous approaches to Visual Question Answering (VQA) have mainly used generic image features from networks trained on the ImageNet dataset, 
despite the wide scope of questions. In contrast, our approach employs features 
derived from networks trained for specialized tasks of scene classification, person activity prediction, and person and 
object attribute prediction.  We also present a method for selecting sub-regions of an image that are relevant for evaluating 
the appropriateness of a putative answer.  Visual features are computed both from the whole image and from local regions, while sentences 
are mapped to a common space using a simple normalized canonical correlation analysis (CCA) model. 
Our results show a significant improvement 
over the previous state of the art, and indicate that answering different question types benefits from examining a variety of 
image cues and carefully choosing informative image sub-regions.

\end{abstract}

\vspace{-.3cm}
\section{Introduction}
\label{sec:introduction}

Visual Question Answering (VQA)~\cite{VQA} has gained popularity in the past year with the release of several new 
datasets~\cite{malinowski14nips,COCOQA,baiduVQA,VQA,zhu2016cvpr, visualgenome, VisualMadlibs}. Recent high-profile approaches 
to VQA combine Convolutional Neural Networks (CNNs) and Long Short-Term Memory (LSTM) networks to generate answers 
for questions about an image~\cite{baiduVQA,malinowski2015ask,DeepCompositionalQA}. Other works integrate attention 
mechanisms for text-guided analysis of images~\cite{zhu2016cvpr,xuARXIV2015,attentionQA,shihCVPR2015}. In contrast 
to these relatively complex methods, simple CNN+Bag-of-Words~\cite{Zhou_etal_2015} and multi-modal Canonical 
Correlation Analysis (CCA)~\cite{VisualMadlibs} have also been shown to achieve good accuracy.

Despite the broad scope of questions and the diversity of proposed solutions for VQA, all the approaches mentioned above use 
image features computed by a CNN pre-trained for image classification on the ImageNet dataset~\cite{ILSVRC15}, 
for example, VGG-Net~\cite{simonyanVGG}. However, it is unlikely that a single network can capture the full range of 
information needed to correctly answer a question. Consider a question about the position of an object in an image: the answer could involve 
information about the overall scene (\eg, it is in the kitchen), other reference objects (on the table), appearance 
(against the blue wall), or even details about people (behind the girl) or activities (in front of the man holding a glass), \etc.  
In order to better understand an image and answer questions about it, it is necessary to leverage rich and detailed global and local information
instead of relying on generic whole-image features.  

\begin{figure}[t!]
	\includegraphics[width=\linewidth]{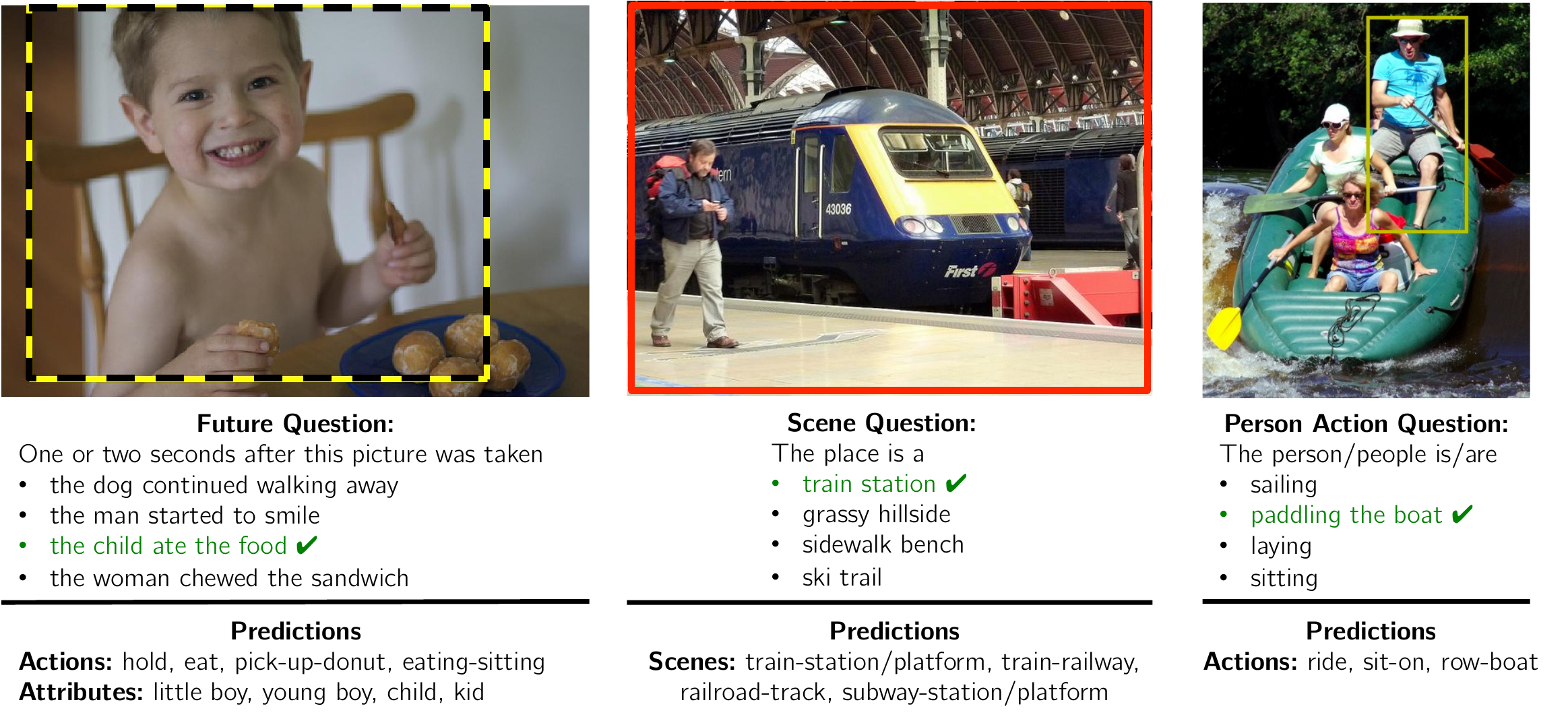}
	\caption{Given Visual Madlibs images and questions, our method uses multiple deep networks trained on external datasets to predict actions, 
	attributes, scenes, and other cues. The spatial support of these cues can be an automatically detected region (left example, dashed yellow box); 
	the whole image (middle example, red box), or a ground truth box provided with the question (right example, yellow box). Top-scoring predicted 
	labels for the corresponding regions are shown on the bottom. We train CCA models on top of cue-specific activations and combine the CCA 
	scores to rank candidate answers for 12 types of multiple choice questions.}
	\label{figure:overview}
\end{figure}

This paper presents a CCA-based approach that uses a variety of cues
to improve performance significantly on questions from the Visual 
Madlibs Dataset~\cite{VisualMadlibs}. This dataset was created by asking people to 
write fill-in-the-blank descriptions for 12 question types, broadly divided into three areas: questions about image as a whole (scene identity, 
emotion evoked by the image, most interesting thing about the image, likely future and past events); questions about an indicated person (attribute, 
action, location, interaction with an indicated object); and questions about an indicated object (affordance, attribute, location). Every Madlibs 
question consists of an image (possibly with a ground truth mask of the indicated person or object), a sentence 
prompt based on the question type, and four possible answers to fill in the blank, one of which is correct. Three example questions are 
illustrated in Figure~\ref{figure:overview}.

A few recent works have similarly attempted to leverage external knowledge, either through a rich set of different labels, or by 
exploiting textual resources such as DBpedia~\cite{dbpedia}. The former approach is adopted in~\cite{KB_feifei_ARXIV} by learning an 
MRF model on scene category, attribute, and affordance labels over images from the SUN dataset~\cite{xiao2010sun}. While quite 
powerful on the image side, the lack of natural language integration limits the set of possible questions that may be asked of
the system. The approach presented in~\cite{CVPR16AMA} starts from multiple labels predicted from images 
and uses them to query Dbpedia. The obtained textual paragraphs are then coded as a feature and 
used to generate answers through an LSTM. Though quite interesting, this 
method still relies on ImageNet-trained features, missing the 
variety of visual cues that can be obtained from networks tuned on tasks other than object classification.
\smallskip

To extract diverse cues for answering Madlibs questions, we use features from CNNs trained  on multiple specialized sources: the Places 
scene dataset~\cite{zhou2014learning},  the HICO and MPII human activity datasets~\cite{HICO,MPII}, 
the MS-COCO object detection dataset~\cite{MSCOCO}, and the grounded image description dataset, Flickr30k Entities~\cite{Flickr30kEntities_JOURNAL}.
As detailed in Section \ref{sec:specific_cnn_models}, our networks are based on state-of-the-art architectures for image 
classification~\cite{simonyanVGG}, object detection~\cite{liu15ssd}, and
action recognition~\cite{ArunARXIV}. Further, we propose methods for automatically finding spatial support for mentions of persons 
and objects in candidate answers (Section \ref{subsec:image_region_selection}) and for combining multiple types of cues to obtain a single 
score for a question and a candidate answer (Section \ref{sec:multi_cue_theory}). In Section \ref{sec:fitb}, we present a thorough experimental 
analysis of different types of cues and combination schemes. By combining information from all cues relevant for a given question type, 
we achieve new state-of-the-art accuracies on all question types.

\section{The Approach}
\label{sec:approach}

We want to learn a model that maps visual information extracted from an image
to the correct multiple-choice answer for a fill-in-the-blank question. 
This image-to-answer compatibility naturally depends on the global compatibility of whole-image cues with the answer sentence, 
as well as on the local compatibility of image regions and any person or object named in the answer sentence. 

Our image-to-text compatibility scores come from CCA models~\cite{gong2014multi}, which give us linear mappings from visual and textual 
features to a common latent space. On the visual side, we leverage deep networks to obtain cue-specific features from the whole image and 
from person/object bounding boxes. We use the activations of the last fully connected layer (fc7) or the final prediction layer 
of each network described in the following section as a visual representation. On the textual side, we represent each word with its 
word2vec feature~\cite{Word2Vec} and consider the average 300-dimensional vector over the whole answer (when picking a choice) or the parsed 
phrases that mention people or objects (when selecting a region of interest). At test time, among the four candidate answers, we select the 
one that obtains the highest cosine similarity with the image features in the joint embedding space of the CCA model. To integrate multiple cues, we
experiment both with low-level visual feature stacking and high-level CCA score combinations.

In the following we provide details about the architectures used to extract visual features (Sec.~\ref{sec:specific_cnn_models}), 
the selection procedure to localize objects and persons named in the sentences (Sec.~\ref{subsec:image_region_selection}), and 
the strategies adopted for multi-cue combination depending on the image question type (Sec.~\ref{sec:multi_cue_theory}).

\vspace{-.1cm}
\subsection{Cue-Specific Models}
\label{sec:specific_cnn_models}
\vspace{-.1cm}

\noindent{\bf Baseline Network (VGG).} 
The VGG-16 network~\cite{simonyanVGG} trained on 1000 ImageNet categories is a standard architecture in many current works on VQA. 
Consistent with the original Madlibs paper~\cite{VisualMadlibs}, we consider it as the reference baseline.  We obtain a 4096-dimensional 
feature vector by averaging fc7 activations over 10 crops.\smallskip

\noindent{\bf Scene Prediction (Places).} The Places dataset~\cite{zhou2014learning} contains about 2.5 million images belonging to 
205 different scene categories. We utilize the VGG-16 network from~\cite{zhou2014learning} with 10-crop 4096 dimensional 
fc7 features to obtain information about the global scene.\smallskip

\noindent{\bf Person Activity Prediction (Act. HICO, MPII).} We leverage two of the largest currently available human activity 
image datasets: the Humans Interacting with Common Objects (HICO) dataset~\cite{HICO} and the  MPII Human Pose Dataset~\cite{MPII}. 
HICO has 600 labels for different human-object interactions, \eg, ride-bicycle or repair-bicycle. The HICO objects belong to the 80 
annotated categories in the MS-COCO dataset~\cite{MSCOCO}. The MPII dataset has 393 categories, including interactions with objects 
as well as solo human activities such as walking and running. We employ the CNN architecture introduced by Mallya and 
Lazebnik~\cite{ArunARXIV}, which currently hold state-of-the-art classification accuracy on both datasets. This architecture is based 
on VGG-16 and combines information from a person bounding box (either ground truth or detected, depending on the question type) and the whole 
image. In case of multiple people in an image, the network is run independently on each person and then the features are average-pooled. 
As will be explained in Section \ref{sec:multi_cue_theory}, for combining multiple cues, we will experiment 
with 4096-dimensional fc7 activations as well as with class prediction logits (inputs to the final sigmoid/softmax layer). \smallskip

\begin{figure}[t!]
	\includegraphics[width=\linewidth]{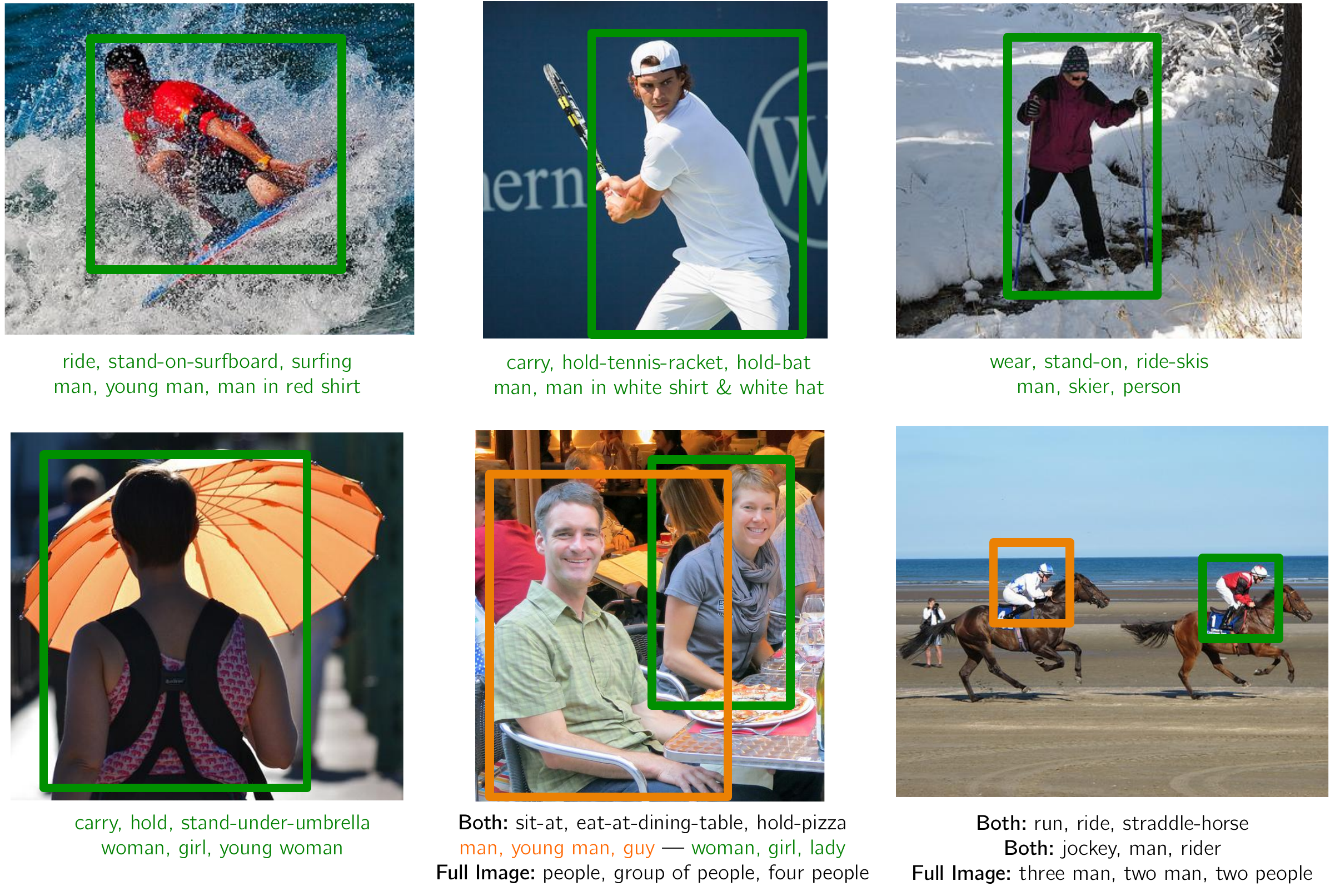}
	\vspace{-5mm}
	\caption{Predicted person actions and attributes. The first and second lines below each image show the top 
	3 predicted actions and attributes respectively. In the case of multiple people in an image, the third line 
	shows the top 3 attributes predicted for the whole image. ({\bf Both} means that both of the people in the 
	image have the same action/attribute predictions.)}
	\label{figure:act_attr_preds}
\end{figure}

\noindent{\bf Person Attribute Prediction (Attr.).} We extract a rich vocabulary of describable person attributes from the Flickr30k Entities 
dataset~\cite{Flickr30kEntities_JOURNAL}, which links phrases in sentences to corresponding bounding boxes in images. Our vocabulary 
consists of 302 phrases that refer to people and occur at least 50 times in the training set, and covers references to gender (man, woman), 
age (baby, elderly man), clothing (man in blue shirt, woman in black dress), appearance (brunette woman, Asian man), 
multiple people (two men, group of people), and more. Besides having the appealing characteristic of being derived 
from natural language phrases, our set of attribute labels is one order of magnitude larger than that of 
other existing attribute datasets~\cite{bourdev2011BAPD,Sudowe2015Parse27k}. 
We train a Fast-RCNN VGG-16 network~\cite{girshick2015fast} to predict our 302 attribute labels based on person bounding boxes 
(in case of group attributes, the ground truth boxes contain multiple people). To compensate for unbalanced training data, we use a 
weighted loss that penalizes mistakes on positive examples 
10 times more than on negative examples~\cite{ArunARXIV}. Unlike our
activity prediction network, this network can predict group attributes given a box containing multiple people. For the downstream 
VQA models, we will consider both the fc7 activations and the class prediction logits of this network, same as with the HICO and MPII networks. 
Sample outputs of the person action and attribute predictors are shown in Figure~\ref{figure:act_attr_preds}. \smallskip

\noindent{\bf Color Prediction (Color).} For questions focused on objects, color is a very salient characteristic that is not captured 
well by CNNs trained to recognize object categories. We follow~\cite{Flickr30kEntities_JOURNAL} and fine-tune a Fast-RCNN VGG-16 network 
to predict one of 11 colors that occur at least 1,000 times in the Flickr30K Entities training set: black, 
red, blue, white, green, yellow, brown, orange, pink, gray, purple. The training is performed on non-person phrases to prevent 
confusion with color terms that refer to race. For VQA, we use the 4096-dimensional fc7 feature extracted from 
the object bounding box.\\ \vspace{-2mm}

\vspace{-.2cm}
\subsection{Image Region Selection}
\label{subsec:image_region_selection}
\vspace{-.1cm}

Some of the Madlibs question types ask about a particular object or person and provide its bounding box
(\eg, the rightmost example in Figure~\ref{figure:overview} asks what the person outlined in yellow is doing). Other questions, namely 
those related to image interestingness, future, and past, do not provide a target image region. In particular, for the left example in 
Figure~\ref{figure:overview}, each of the four candidate answers mentions a different person or object: ``the dog,'' ``the man,'' ``the child,'' 
``the woman.'' In order to pick the right choice, we need to select the best supporting regions for each of the four entity mentions and use 
the respective correspondence scores as part of our overall image-to-answer scoring scheme. 

For Interestingness, Past, and Future questions, we first parse all answers with the Stanford parser \citep{SocherEtAl2013:CVG} and use pre-defined 
vocabularies to identify NP (Noun-Phrase) chunks referring to a person or to an object. Then we apply the following region selection mechanisms for 
mentioned people and objects, respectively.\smallskip

\begin{figure}[t!]
	\includegraphics[width=\linewidth]{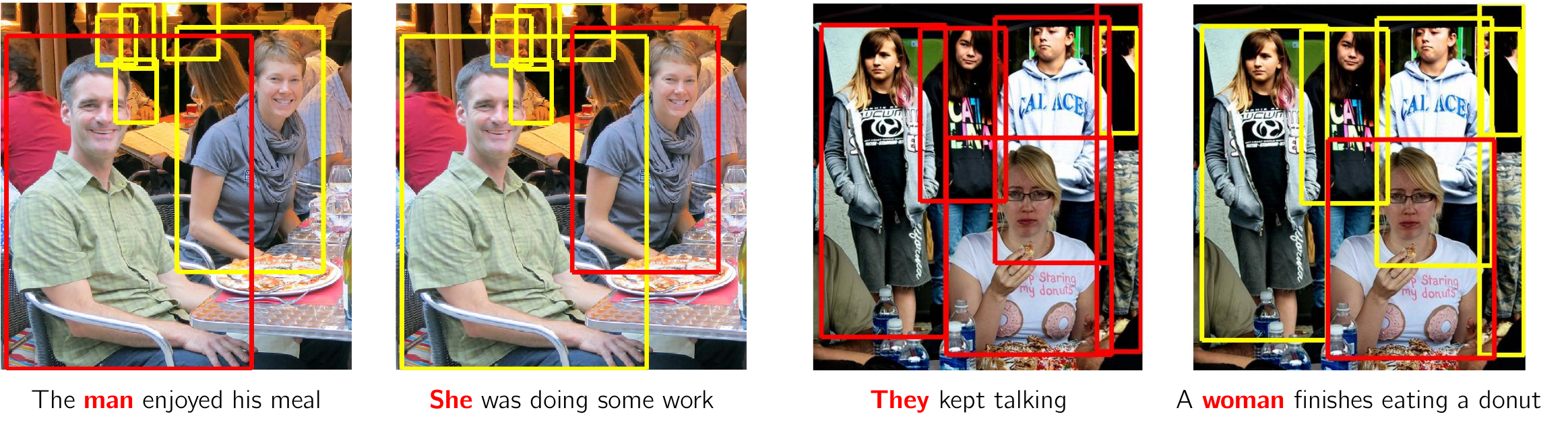}
	\vspace{-5mm}
	\caption{Examples of selected person boxes based on person phrases (in red).}
	\label{figure:box_selection}
\end{figure}

\noindent{\bf Person Box.} We first detect people in an image using the Faster-RCNN detector~\cite{ren2015faster} with the default confidence 
threshold of 0.8. We discard all detected boxes with height or width less than 50 
pixels since in our experience they mostly contain noise and fragments. We also consider 
the smallest box containing all detected people, to account for cues originating from multiple people. 
Given an image and an answer, we attempt to select the person detection that corresponds best to the named person. For example, 
if an answer refers to a ``young girl,'' we want to select the detection window that looks the most like a young girl.
To this end, we train a {\bf Person CCA model} on the val+test set of Flickr30k Entities using person phrases and 
image features extracted from the corresponding person box. We represent the phrases with the 300-d average of word2vec~\cite{Word2Vec} and the 
image regions with the 302-d vector of predictions obtained from our person attribute network 
(Sec.~\ref{sec:specific_cnn_models}). To apply this model to the Madlibs dataset, we extract the part 
of the answer sentence referring to a person and select the image region with the highest similarity in the CCA embedding space.  A few successful 
region selections are shown in Figure~\ref{figure:box_selection} (parsed person phrase and corresponding selected boxes are colored red). 
Note that in the third example, CCA selects the overall box. Thus, all the person-specific boxes are colored red with the exception of the 
top right one which is discarded as it is below the size threshold. In case no words referring to people are found in a choice, all person boxes 
are selected.\smallskip

\begin{figure}[t!]
	\includegraphics[width=\linewidth]{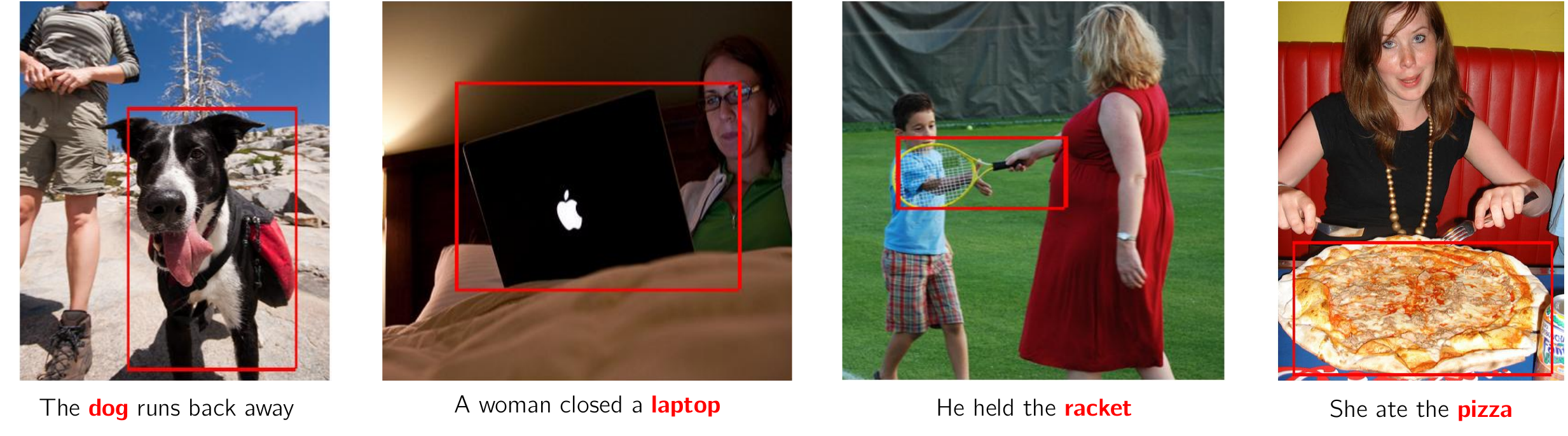}
	\caption{Examples of selected object boxes based on object phrases (in red).}
	\label{figure:obj_box_selection}
\end{figure}

\noindent{\bf Object Box.}
We localize objects using the Single Shot MultiBox Detector (SSD)~\cite{liu15ssd} that has been pre-trained on the 
80 MS-COCO object categories. SSD is currently state-of-the-art for detection in speed and accuracy. 
For each Madlibs image, we consider the top 200 detections as object candidates and use the {\bf Object CCA model}
created for the phrase localization approach of~\cite{Flickr30kEntities_JOURNAL} 
to select the boxes corresponding to objects named in the sentences.
This model is trained 
on the Flickr30k Entities dataset over Fast-RCNN fc7 features and 300-d word2vec features.
The top-scoring box is selected for each object phrase (Figure~\ref{figure:obj_box_selection}). \smallskip

\noindent{\bf Person and Object Scores.} 
The Person and Object CCA models created for image region selection can also be used to help score 
multiple-choice answers. For the detected people, we associate each answer with the score of the 
selected person box. For the objects, since the detector output is much less reliable and 
the cues are more noisy, we use a kernel introduced for matching sets of local features~\cite{Lyu:05}: we collect all of the $N$ object 
boxes from the image and the $M$ object phrases from each answer and then compute the following score:
\begin{equation}
K(image, answer) = \frac{1}{N}\frac{1}{M}\sum_{i=1}^N\sum_{j=1}^M \{\text{cos\_similarity}(box_i, phrase_j)\}^p~.
\vspace{-2mm}
\end{equation}
The parameter $p$ assigns more relative weight to box-phrase pairs with higher similarity. We use $p=5$ in our implementation. 

\subsection{Cue Integration}
\label{sec:multi_cue_theory}
With the variety of features described in section  \ref{sec:specific_cnn_models} we can cover different
visual aspects of the image that are relevant for different question types.
For each Madlibs image, we extract the global VGG and Places representations and use either the ground 
truth focus region (if provided with the question type) or the region selection procedure of Section 
\ref{subsec:image_region_selection} to localize where VGG, Act., Attr., and Color features should be 
extracted for persons and objects. 
As detailed below, we then explore a variety of combination schemes, including stacking of various network outputs and forming 
linear combinations of scores from CCA models learned on top of different features. 
\smallskip

\noindent{\bf Combining fc7 activations.} Our most basic integration scheme is to combine the output of the vanilla global VGG network 
with the output of a single cue-specific network applied either to the whole image (Places), or to a person or object bounding box. 
To do this, we stack the 4096-d fc7 activations of the respective networks to obtain 8192-d representations. Results with this scheme 
will be shown in the ``Baseline + X'' columns of Table 
 \ref{table:madlibs_accuracy}. \smallskip
 
 \noindent {\bf Combining label activations.} If we want to combine the VGG baseline with the output of more than one specialized network, 
 fc7 stacking becomes infeasible due to excessively high dimensionality. Instead, we stack lower-dimensional class prediction vectors 
 (recall from Section \ref{sec:specific_cnn_models} that we use logits, or values before the final sigmoid/softmax layer). Specifically, 
 to characterize people, we concatenate the class predictions of the two action networks (HICO+MPII), or the two action networks together 
 with the attribute network (HICO+MPII+Attr.), giving 993-d and 1295-d feature vectors, respectively. Results with this scheme will be 
 shown in the Label Combination columns of Table \ref{table:multi_cue}. \smallskip

\noindent{\bf CCA Score Combination.} To enable even more complex cue integration, we resort to combining scores of multiple CCA models. 
Namely, for each of the stacked cues described above, we learn a CCA model on the training portion of the Madlibs dataset. Given a test 
question/answer pair, we obtain the score of each CCA model that is appropriate for that question type and linearly combine the scores with 
weights chosen depending on the question type. 
From the $C$ available cues for that type, we manually pre-determine the one that makes the most sense (\eg, Places for person location, 
Color for object attribute) and assign it a weight of $(1- (C-1)\times0.1)$ 
while all of the remaining cues get weight $0.1$. Once the weighted CCA score is calculated for 
all the candidate answers, we choose the one with the highest score. The resulting performance will be shown 
in the last three columns of Table \ref{table:multi_cue}.

\section{Experiments}
\label{sec:experiments}
\label{sec:fitb}

\begin{table}[tb!]
	\setlength{\tabcolsep}{5pt}
	\footnotesize
	\begin{center}
		\begin{tabular}{@{}lll|c||c|c|c|c|c|c|c@{}} 
            \hline
            & \multicolumn{2}{c|}{\multirow{2}{*}{Question Type}} & \multicolumn{2}{c|}{Full Image} & \multicolumn{4}{c|}{Person Box} & \multicolumn{2}{c}{Object Box}\\ 
            \cline{4-11}
            & & & Baseline & B. +  & B. + & B. + & B. + & B. + & B. + & B. +\\
            & & & VGG      & Places      & VGG        & Act. HICO  & Act. MPII  & Attr.      & VGG        & Color\\\hline
            \multirow{10}{*}{a)} & \multirow{2}{*}{Scene} & Easy & 88.14 & \bf{89.48} & -- & -- & -- & -- & -- & --\\ 
										& & Hard & 71.05 & \bf{73.42} & -- & -- & -- & -- & -- & --\\ 
			& \multirow{2}{*}{Emotion}	& Easy & 52.84 & \bf{52.92} & -- & -- & -- & -- & -- & --\\ 
										& & Hard & \bf{40.07} & 39.72 & -- & -- & -- & -- & -- & --\\ 
			& \multirow{2}{*}{Interesting} & Easy & 79.53 & 79.74 & 79.92 & \bf{80.82} & 80.51 & 80.04 & 79.61 & --\\ 
										&  & Hard & 55.05 & 55.05 & 55.45 & 54.95 & 55.83 & \bf{55.99} & 54.92 & --\\ 
			& \multirow{2}{*}{Past}		& Easy & 80.24 & 80.86 & 81.27 & \bf{83.09}& 81.56	& 82.68 & 80.75 & --\\ 
										& & Hard & 54.35 & 54.64 & 55.74 & 55.61 & 55.57& \bf{57.74}& 54.82 & --\\ 
			& \multirow{2}{*}{Future}   & Easy & 80.22 & 80.96 & 81.47 & 82.84	& 81.62  & \bf{83.19} & 81.79 & --\\ 
										& & Hard & 55.49 & 56.03 & 57.51 & 57.36 & 56.72  & \bf{59.21} & 57.26 & --\\ \hline
			\multirow{8}{*}{b)} & Person & Easy & 53.56 & 54.50 & 60.04 & 54.86 & 55.66	& {\bf 64.97} & -- & -- \\ 
					& \quad Attribute & Hard & 42.58 & 42.89 & 49.34 & 43.79 & 45.85	& {\bf 55.50} & -- & --\\ 
            		& Person  	& Easy & 84.71 & 84.89 & 85.96 & {\bf 87.54} & 85.46 & 85.13 & --& --\\ 
            		& \quad Action  	& Hard & 68.04 & 68.68 & 69.79 & {\bf 71.39} & 70.33 & 69.08 & --& --\\ 
            		& Person  	& Easy & 84.95 & {\bf 86.16} & 84.70 & 85.49 & 85.12 & 84.48 & --& --\\ 
            		&\quad Location  	& Hard & 64.67 & {\bf 66.72} & 65.50 & 64.91 & 65.36 & 64.77 & --& --\\ 
					& Person Object & Easy & 73.63 & 74.52 & 75.26 & {\bf 78.34} & 76.66 & 75.59 & 77.06 & 75.84\\
					&\quad Relationship & Hard & 56.19 & 56.88 & 59.06 & {\bf 60.37} & 59.27 & 58.35 & 57.17 & 57.45 \\ \hline
			\multirow{6}{*}{c)} & Object & Easy & 50.35 & 50.64	 & -- & -- & -- & -- & 57.56 & \bf{59.31}	\\ 
					& \quad Attribute	 & Hard & 45.41 & 45.55	 & -- & -- & -- & -- & 53.63 & \bf{54.73}	\\ 
					& Object		& Easy & 82.49 & 83.10	 & -- & -- & -- & -- & \bf{87.40} & 84.02	\\ 
					& \quad Affordance  & Hard & 64.46 & 64.55	 & -- & -- & -- & -- & \bf{68.47} & 65.37	\\ 
                    & Object		& Easy & 67.91 & \bf{69.75} & -- & -- & -- & --	& 68.68	& 69.22	\\ 
					& \quad Location 	& Hard & 56.71 & \bf{58.08} & -- & -- & -- & --	& 57.90	& 57.35	\\ 
        \end{tabular}\vspace{-2mm}
	\end{center}
	\caption{Accuracy on Madlibs questions with fc7 features. The Baseline VGG column gives performance for 4096-d fc7 
    outputs of the standard reference network trained on Imagenet. For the columns labeled ``B. + X,'' the baseline fc7 features are concatenated with 
    fc7 features of different specialized networks, yielding 8192-d representations (see Section \ref{sec:multi_cue_theory}). }
	\label{table:madlibs_accuracy}
\end{table}

As mentioned earlier, the 12 types of Madlibs questions can be broadly divided into three groups based on whether they are about
the whole image, a specific person, or a specific object. In the first group there are questions related to scene,
emotion, interestingness, past, and future. The second group asks questions about specified people, including attributes, activities, location, 
and relationship with an object. The third group asks questions about attributes, affordances, and position of a specified object. ``Hard'' and 
``Easy'' versions of each question are provided with the dataset (``Hard'' questions have distractor options that are more easily confused with 
the correct option). To start, the leftmost column
of Table~\ref{table:madlibs_accuracy} presents accuracies for each question type when using the baseline global VGG feature,
while the following columns show the performance for feature combination of the baseline with the individual cues. 
We want to see how using cues better suited for different question types can improve performance.
\smallskip

\noindent{\bf Whole-Image Questions.} As shown in group (a) of Table~\ref{table:madlibs_accuracy}, for Scene questions, using the fc7 Places 
features helps improve performance over the VGG baseline. Emotion questions are rather difficult to answer and we do not see much improvement 
by adding scene-based features. We did not attempt to use person- or object-based features for the Scene and Emotion questions since we found 
that only 13\% (resp. 2\%) of the answers to those two question types mention one of the 80 MS-COCO objects and less than 2\% mention one of the 
302 person labels.

On the other hand, for the Future, Past, and Interestingness questions, people and objects often play an important role: 
between 30\% and 40\% of the answers name an object and the frequency of person mentions ranges from 25\% for Interestingness 
to about 80\% for Past and Future. Thus, for these question types, we perform person and object detection and use the selection methods 
described in Sec.~\ref{subsec:image_region_selection} to find relevant boxes to score a given answer. We extract 
four different types of fc7 features from a selected person box: VGG features from passing a resized box 
($224\times 224$) as input, Act. features from the networks trained on HICO and MPII, and the Attr.
features from the prediction network trained on Flickr30. 
We do not expect color to provide useful information to discriminate between answers, so we do not include it here.
From Table~\ref{table:madlibs_accuracy} (a), we find that Act. and Attr. features give us improvement in accuracy 
with respect to the whole image baseline. The HICO network, with its large number of labels covering 
objects from the MS-COCO dataset, provides better results than the MPII network. 
However, VGG features extracted from the object regions do not help to improve over the whole-image baseline.
\smallskip

\noindent{\bf Person Questions.}
For questions about specified people, we extract features from the provided ground truth person box and report results in 
group (b) of Table~\ref{table:madlibs_accuracy}. As expected, attribute features yield the best results on 
Attribute questions and the HICO representation improves accuracy by up to 
3\% over the baseline for Action questions. For Person Location, the most useful representation 
is the one obtained from the Places dataset. Finally, for the Person-Object Relation questions, 51\% of answers name one of the 600 
HICO actions, explaining the observed performance boost obtained with HICO. For the latter question type, the ground truth position of the 
query object is also provided: by extracting the VGG and Color features from the object box we obtain lower accuracies than with the HICO 
representation but higher than with the whole-image baseline.
\smallskip


\noindent{\bf Object Questions.}
For questions about specified objects, we extract features from the provided ground truth object box and report results in group (c) of 
Table~\ref{table:madlibs_accuracy}. Here, 
the best results for Attribute questions are obtained with the Color representation, the best results for Affordance
questions are obtained with the VGG representation, and the best results for Object Location are obtained
with the Places representation. 
\smallskip

\begin{table}[tb!]
	\setlength{\tabcolsep}{2pt}
	\footnotesize
	\begin{center}
		\begin{tabular}{lll||c|cc||c|c||c|c||c}
		\hline
& \multicolumn{2}{c||}{\multirow{3}{*}{Question Type}} & \multicolumn{3}{|c||}{fc7 Combination } & \multicolumn{2}{|c||}{Label Combination} & \multicolumn{3}{|c}{CCA Score Combination} \\
\cline{4-11} 
& & & Baseline & \multicolumn{2}{c||}{Baseline +} & HICO & HICO + MPII & + Person & + Obj. & CCA \\
& & & VGG & \multicolumn{2}{c||}{ Single Best Cue} & + MPII & + Attr. & Score & Score & Ensemble\\\hline
\multirow{6}{*}{a)}& \multirow{2}{*}{Interesting} & Easy & 79.53 & HICO & 80.82 &79.96 & 81.12 & 81.69 & 81.57 & {\bf 83.20}\\ 
							& & Hard & 55.05 & Attr. & 55.99 &53.95 & 55.76 & 56.64 & 56.37 & {\bf 57.70}\\
& \multirow{2}{*}{Past}		  & Easy & 80.24 & HICO & 83.09 &83.29 & 84.64 & 85.62 & 85.05 & {\bf 86.36}\\ 
                            & & Hard & 54.35 & Attr. & 57.74 &55.23 & 58.21 & {\bf 60.33} & 58.43 & 60.00\\
& \multirow{2}{*}{Future}	  & Easy & 80.22 & Attr. & 83.19 &83.66 & 85.53 & 85.79 & 85.57 & {\bf 86.88}\\ 
							& & Hard & 55.49 & Attr. & 59.21 &57.58 & 60.61 & 61.85 & 60.63 & {\bf 62.39}\\\hline
\multirow{8}{*}{b)}& Person & Easy & 53.56 & Attr. & 64.97 &60.22 & 67.96 & -- &-- & {\bf 68.50}\\ 
				& \quad Attribute & Hard & 42.58 & Attr. & 55.50 &46.44 & 55.78 & -- &-- & {\bf 55.90}\\
				&	Person & Easy & 84.71 & HICO & 87.54 &87.31 & 87.56 & -- &-- & {\bf 88.34}\\ 
				&	\quad Action & Hard & 68.04 & HICO & 71.39 &71.16 & 71.56 & -- &-- & {\bf 71.65}\\   
				&	Person & Easy & 84.95 & Places & {\bf 86.16} &84.77 & 84.80 & -- &-- & 85.70\\ 
				&	\quad Location & Hard & 64.67 & Places & {\bf 66.72} &62.65 & 62.80 & -- &-- & 63.92\\
				&	Person Object & Easy & 73.63 & HICO & 78.34 &77.49 & 77.77 & -- &-- & {\bf 78.93}\\ 
				&	\quad Relationship & Hard & 56.19 & HICO & {\bf 60.37} &57.91 & 57.96 & -- &-- & 58.63\\\hline
\multirow{6}{*}{c)}& Object    & Easy & 50.35 & Color & {\bf 59.31} &-- &-- &-- &-- & 58.94\\
        & \quad Attribute & Hard & 45.41 & Color & {\bf 54.73} &-- &-- &-- &-- & 54.50\\
		& Object	   & Easy & 82.49 & Obj.\ VGG & {\bf 87.40} &-- &-- &-- &-- & 87.29\\
        & \quad Affordance & Hard & 64.46 & Obj.\ VGG & {\bf 68.47} &-- &-- &-- &-- & 68.37\\             
        & Object   & Easy & 67.91 & Places & 69.75 &-- &-- &-- &-- &{\bf 70.03}\\
        & \quad Location & Hard & 56.71 & Places & {\bf 58.08} &-- &-- &-- &-- & 58.01\\                      
        \end{tabular}
	\end{center} 
    \caption{Results of combining multiple cues. Columns marked ``fc7 Combination'' give key results from Table \ref{table:madlibs_accuracy} for 
    reference. Columns marked ``Label Combination'' show results with the respectively named strategies of Section \ref{sec:multi_cue_theory}. 
    Columns marked ``+ Person Score'' and ``+ Obj. Score'' show the results of combining the region selection scores of 
    Section \ref{subsec:image_region_selection} with the HICO + MPII + Attr. CCA. The CCA Ensemble column shows the results of linearly combining 
    all CCA scores appropriate for each question type (see text for details).}
	\label{table:multi_cue}
\end{table}

\noindent{\bf Cue Integration.} Table~\ref{table:multi_cue} reports results obtained by integrating multiple cues as described in Section 
\ref{sec:multi_cue_theory}. We exclude Scene and Emotion questions 
from the following analysis since they do not involve persons and objects and we previously only used a single 
cue for them. Second and third columns reproduce the best results 
on the different question types, as previously reported in Table~\ref{table:madlibs_accuracy}, and the subsequent columns compare
performance obtained by integrating multiple cues via label combination and CCA score.

For questions in groups (a) and (b) of Table~\ref{table:multi_cue}, we test the representations obtained by action (HICO+MPII) and 
attribute (HICO+MPII+Attr.) label combination. For HICO+MPII, we typically see a small drop in 
performance on whole-image questions (\ie, in Interesting, Past, Future rows) and location-related questions 
(Person Location and Person-Object Relation), probably owing to the reduced feature dimension and loss of global contextual information 
as compared to the 8192-dimensional fc7 combination features.  
On the other hand, the HICO+MPII representation produces results comparable with the best fc7 cue for the Person Action question 
while being much more compact (993-d vs. 8192-d). By adding the attribute labels (HICO+MPII+Attr. column), we further improve performance, 
particularly for the Person Attribute question.

The last three columns of Table \ref{table:multi_cue} shows the results of linearly combining multiple CCA scores 
as described in the last paragraph of Section \ref{sec:multi_cue_theory}. 

Recall from Section \ref{subsec:image_region_selection} that for Interestingness, Past, and Future questions, we perform focus region 
selection and compute Person and Object scores measuring the compatibility of person and object mentions in answers with the selected regions. 
These scores also provide some useful signal for choosing the correct answer, so we combine them (with weight 0.1) with the scores from the 
HICO+MPII+Attr. column (with weight 0.9).
The results reported in columns labeled ``+ Person Score'' and ``+ Obj. Score'' of Table~\ref{table:multi_cue} show 
small but consistent accuracy improvements, particularly for the hard questions.

Finally, we consider different CCA score ensemble depending on the question type.
For Interestingness, Past, and Future questions we combine scores from CCA models trained on Places, 
VGG Person Box, and VGG Object Box features (as in Table~\ref{table:madlibs_accuracy}),
with the ``+ Person Score'' from region selection. 
For Person questions, we combine CCA scores from Places, VGG Person Box, and 
HICO+MPII+Attr. models. For Object questions, we combine CCA scores from Places, 
VGG Object Box, and Color models. 
Overall, we observe an average improvement of about 1\% in accuracy for most of the questions with respect to the single 
best cue performance and 5\% with respect to the baseline.

\section{Conclusions}
\label{sec:conclusion}
We have shown that features representing different facets of image content are helpful for answering 
multiple choice questions. This indicates that external knowledge can be successfully transferred with 
the help of deep networks trained on specialized datasets. Further, by attempting to match image regions 
with the persons/objects named in the answers, and by using 
an ensemble of CCA models, we have created a system that beats the previous state of the art on Visual Madlibs 
and establishes a stronger set of baselines for future methods to beat. As future work, besides evaluating the 
proposed approach with other multiple-choice question datasets and searching for other sources of external knowledge, 
we plan to improve our multi-cue integration method by learning combination weights for each feature.

\smallskip
\noindent{\bf Acknowledgments.} This material is based upon work supported by the National Science Foundation under 
grants 1302438, 1563727, 1405822, 1444234, Xerox UAC, Microsoft Research Faculty Fellowship, and the Sloan Foundation Fellowship.

\bibliography{egbib}

\end{document}